\documentclass[letterpaper, 10 pt, conference]{ieeeconf}  

\IEEEoverridecommandlockouts                              

\usepackage{threeparttable}
\let\labelindent\relax
\usepackage{enumitem}
\usepackage{booktabs}
\usepackage{pgfplots}
\usepackage{graphicx}
\usepackage{colortbl}
\usepackage{subcaption}
\usepackage{makecell}
\usepackage{multirow}
\usepackage{bm}
\usepackage{algpseudocode}
\usepackage{algorithm}
\usepackage{tabularx}
\usetikzlibrary{plotmarks}
\usepackage{url}
\usepackage{tikz}

\usepackage{amsmath}
\usepackage{cite}
\usepackage[colorlinks=true,linkcolor=black,citecolor=black,urlcolor=black]{hyperref}
\definecolor{lightgreen}{rgb}{0.9, 1.0, 0.9}

\newcommand\markersize{2.5pt}

\newcommand\linesize{3.0pt}
\newcommand{\axisfont}{\fontsize{15}{18}\selectfont} 

\definecolor{dense}{rgb}{0.6350 0.0780 0.1840}
\definecolor{sparse}{rgb}{0.4660 0.6740 0.1880}
\definecolor{moderate}{rgb}{0.3010 0.7450 0.9330}

\definecolor{sourcecolor}{rgb}{0.85000,0.32500,0.09800}%
\definecolor{targetcolor}{rgb}{0.00000,0.44700,0.74100}%
\definecolor{rotatedcolor}{rgb}{0.92900,0.69400,0.12500}%
\newcommand{\pointsource}{%
	\tikz[baseline=-0.6ex]{%
		\draw [fill=sourcecolor, sourcecolor]
		(0,0) circle (3pt);%
	}%
}

\newcommand{\pointtarget}{%
	\tikz[baseline=-0.6ex]{%
		\draw [fill=targetcolor, targetcolor]
		(0,0) circle (3pt);%
	}%
}

\newcommand{\pointrotated}{%
	\tikz[baseline=-0.6ex]{%
		\draw [fill=rotatedcolor, rotatedcolor]
		(0,0) circle (3pt);%
	}%
}

\newcommand{\pointexcluded}{%
	\tikz[baseline=-0.6ex]{%
		\draw [fill=black, black]
		(0,0) circle (3pt);%
	}%
}

\newcommand{\bboxline}{%
	\tikz[baseline=-0.6ex]{%
		\draw [green, ultra thick]
		(0,0) -- (0.3,0);%
	}%
}

\newcommand{\arrowUpModerate}{%
  \tikz[baseline=-0.15ex]{%
    \draw[->, >=stealth, ultra thick, line width=2pt, moderate]
      (0,-0.1) -- (0,0.3);
  }%
}

\newcommand{\arrowUpDense}{%
  \tikz[baseline=-0.15ex]{%
    \draw[->, >=stealth, ultra thick, line width=2pt, dense]
      (0,-0.1) -- (0,0.3);
  }%
}
\title{\LARGE \bf
Class-Aware PillarMix: Can Mixed Sample Data Augmentation Enhance 3D Object Detection with Radar Point Clouds?
}

\author{Miao Zhang$^{1,2}$, Sherif Abdulatif$^{1}$, Benedikt Loesch$^{1}$, Marco Altmann$^{1}$ and Bin Yang$^{2}$
\thanks{$^{1}$Robert Bosch GmbH, Corporate Research, 71272, Germany.
		{\tt \{miao.zhang5, sherif.abdulatif, benedikt.loesch, marco.altmann\}@de.bosch.com}}%
	\thanks{$^{2}$University of Stuttgart, Institute of Signal Processing and System Theory, 70569, Germany.
		{\tt bin.yang@iss.uni-stuttgart.de}}%
}

\begin{document}
\maketitle
\thispagestyle{empty}
\pagestyle{empty}

\begin{abstract}
Due to the significant effort required for data collection and annotation in 3D perception tasks, mixed sample data augmentation (MSDA) has been widely studied to generate diverse training samples by mixing existing data. Among these methods, MixUp is a prominent approach that generates new samples by linearly combining two existing ones, using a mix ratio sampled from a $\beta$ distribution. This simple yet powerful method has inspired numerous variations and applications in 2D and 3D data domains. Recently, many MSDA techniques have been developed for point clouds, but they mainly target LiDAR data, leaving their application to radar point clouds largely unexplored. In this paper, we examine the feasibility of applying existing MSDA methods to radar point clouds and identify several challenges in adapting these techniques. These obstacles stem from the radar's irregular angular distribution, deviations from a single-sensor polar layout in multi-radar setups, and point sparsity. To address these issues, we propose \textit{Class-Aware PillarMix (CAPMix)}, a novel MSDA approach that applies MixUp at the pillar level in 3D point clouds, guided by class labels. Unlike methods that rely a single mix ratio to the entire sample, CAPMix assigns an independent ratio to each pillar, boosting sample diversity. To account for the density of different classes, we use class-specific distributions: for dense objects (e.g., large vehicles), we skew ratios to favor points from another sample, while for sparse objects (e.g., pedestrians), we sample more points from the original. This class-aware mixing retains critical details and enriches each sample with new information, ultimately generating more diverse training data.
Experimental results demonstrate that our method not only significantly boosts performance but also outperforms existing MSDA approaches across two datasets (Bosch Street and K-Radar). We believe that this straightforward yet effective approach will spark further investigation into MSDA techniques for radar data. Code is available at \url{https://github.com/boschresearch/CAPMIX}.

\end{abstract}

\section{INTRODUCTION}
 Autonomous driving technologies have advanced rapidly, driven by significant progress in deep learning. Yet, 3D perception tasks remain highly data-dependent, typically requiring extensively annotated datasets for reliable performance. Collecting and labeling such data in real-world driving scenarios can be highly time-consuming and labor-intensive. To tackle these challenges, mixed sample data augmentation (MSDA) is widely studied~\cite{park2022unified,kong2023lasermix,wang2023ssda3d,xiao2022polarmix,wu2024pillarmix}. By blending LiDAR point clouds from different samples, MSDA effectively expands and diversifies training sets, resulting in notable gains in fully supervised, semi-supervised, and domain adaptation contexts, particularly when data and/or labels are scarce.
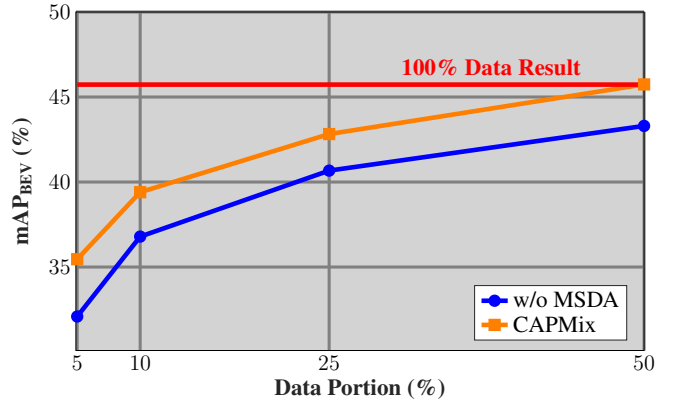
\begin{figure}[tb]
    \vspace{0.5em}
    \centering
    {\resizebox{\columnwidth}{!}{
%
%
\definecolor{mycolor1}{rgb}{0.82745,0.82745,0.82745}%
\begin{tikzpicture}
\pgfplotsset{
		every axis/.append style={
			font=\axisfont 
		}
	}
\begin{axis}[%
width=5.119in,
height=3.05in,
at={(0.603in,0.54in)},
scale only axis,
axis lines=box,
xmin=4.9,
xmax=50,
xtick={ 5, 10, 25, 50},
xlabel style={font=\bfseries\color{white!15!black}\axisfont},
xlabel={Data Portion (\%)},
ymin=30.09,
ymax=50,
ytick={30, 35, 40, 45, 50},
ylabel style={font=\bfseries\color{white!15!black}\axisfont},
ylabel={$\text{mAP}_{\text{BEV}}\text{ (\%)}$},
axis background/.style={fill=mycolor1},
xtick pos=left,
ytick pos=left,
xmajorgrids,
ymajorgrids,
major grid style={line width=2pt, draw=black!50},
legend style={at={(0.97,0.03)}, anchor=south east, legend cell align=left, align=left, draw=white!15!black, font=\axisfont}
]
\addplot [color=blue, line width=\linesize, mark=*, mark options={solid, fill=blue, blue}, mark size=\markersize]
  table[row sep=crcr]{%
5	32.09\\
10	36.79\\
25	40.67\\
50	43.3\\
};
\addlegendentry{w/o MSDA}

\addplot [color=orange, line width=\linesize, mark=square*, mark options={solid, fill=orange, orange}, mark size=\markersize]
  table[row sep=crcr]{%
5	35.46\\
10	39.4\\
25	42.82\\
50	45.74\\
};
\addlegendentry{CAPMix}

\addplot [color=red, line width=\linesize, forget plot]
  table[row sep=crcr]{%
5	45.73\\
50	45.73\\
};
\node[left, align=right, inner sep=0, font=\bfseries\color{red}\axisfont]
at (axis cs:45,46.73) {100\% Data Result};

\end{axis}
\end{tikzpicture}%
}}
    \caption{Comparison between CAPMix and no MSDA reveals remarkable improvements on the Bosch Street dataset.}
    \label{fig:leading}
    \vspace{-2em}
\end{figure}

 Radar sensors transmit electromagnetic waves to measure the range, angle, and velocity of surrounding targets. Their robustness in adverse weather conditions and low cost made them increasingly important for 3D perception tasks in autonomous driving~\cite{srivastav2023radars}.
Compared to LiDARs, radar point clouds are typically sparser and noisier, which intensifies the data scarcity problem. Although recent research has focused on designing advanced models to improve radar perception~\cite{liu2023smurf,ulrich2022improved,musiat2024radarpillars,palffy2022vod}, their performance in low-data scenarios remains relatively limited and largely unexplored.  Since both radar and LiDAR point clouds share a similar format (3D coordinates plus additional attributes), we hypothesize that MSDA methods are also well-suited for radar 3D perception tasks. To investigate this, we assess how state-of-the-art MSDA techniques, originally designed for LiDAR, translate to radar data. Our findings reveal several obstacles, including the irregular angular distribution from radar’s radiation pattern, inherent point sparsity, and deviations from a single-sensor polar distribution in multi-radar setups, all of which limit the direct application of existing MSDA approaches.

In this paper, we propose \textbf{Class-Aware PillarMix (CAPMix)}, a novel approach that mixes radar point clouds with independently determined mix ratios based on class information:
We adopt pillars (columnar grid cells in 3D space) as the operational units for MSDA, as they provide an effective representation of the 3D space. Unlike PillarMix~\cite{wu2024pillarmix}, which directly swaps points, we apply MixUp~\cite{zhang2017mixup} within the pillars to achieve class-aware data augmentation. Similar to MixUp~\cite{zhang2017mixup}, our method samples a mix ratio from a specified $\beta$ distribution to determine the proportion of points taken from the two inputs. To improve sample diversity, each pillar is assigned its own mix ratio. Leveraging class information from the labels, we apply distinct $\beta$ distributions to different pillars for ratio sampling. This strategy ensures that the generated samples are augmented while retaining crucial object-level details.

 Experimental results demonstrate the effectiveness of our method. 
As Fig.~\ref{fig:leading} shows, only using 25\% of the data achieves performance comparable to 50\% of the data without MSDA, while 50\% of the data with CAPMix matches the performance of
the full dataset. This indicates the feasibility and potential of our approach for effectively augmenting radar-based 3D perception under limited data conditions. Our key contributions are summarized below:
\begin{itemize}
    \item  We investigate the applicability of MSDA to radar point clouds and highlight the limitations of existing methods through analysis and experiments.
    \item We propose CAPMix, a straightforward yet effective mixing strategy that uses class information to generate diverse samples for radar point clouds.
    \item We validate CAPMix on two radar datasets (Bosch Street~\cite{armanious2024bosch} and K-Radar~\cite{paek2022k}) of different scales, showcasing its superior performance over other state-of-the-art techniques in terms of $AP_{BEV}$ and $AP_{3D}$.
\end{itemize}
\section{Related Works \label{sec: motivation}}
MSDA generates new
samples by mixing the existing training samples. It was first introduced for image tasks in MixUp~\cite{zhang2017mixup}, which combines input pixels and output labels in a specified ratio. Consequently, it has become an effective technique to enhance model robustness and is compatible with traditional augmentations, such as random rotation and scaling, while also generating more diverse samples for training. Building on this approach, CutMix~\cite{yun2019cutmix} randomly crops a rectangular region from one image and pastes it onto another. More recently, PuzzleMix~\cite{kim2020puzzle} incorporates saliency information from the backward gradient into the mixing process, while AutoMix~\cite{liu2022automix} trains a learnable MixUp generator under the supervision of mixed labels. These pixel-level mixing methods have attracted increasing attention and demonstrate both effectiveness and generalizability in 2D image perception tasks~\cite{naveed2024mixsurvey}. However, fundamental differences between 2D images and 3D point clouds prevent most of these approaches from being directly applied to 3D data.

 Recent works have extended MSDA to 3D perception tasks~\cite{kong2023lasermix,wu2024pillarmix,xiao2022polarmix,wang2023ssda3d} to enhance 3D perception performance across various tasks, such as object detection and semantic segmentation. Given two point clouds $s^A$, $s^B$ with their labels $y^A$ and $y^B$, these methods select points from each of the samples to generate a new sample $\tilde{s}$ and its corresponding label $\tilde{y}$. For example, MixUp~\cite{wang2023ssda3d} randomly selects points by a given mix ratio. PolarMix~\cite{xiao2022polarmix} selects one area from each of the two point clouds based on azimuth angle. Similarly, LaserMix~\cite{kong2023lasermix} partitions the point clouds into multiple regions based on elevation angle, then merges these regions from both point clouds in an alternating manner. Furthermore, PillarMix~\cite{wu2024pillarmix} performs alternating selection on pillars instead of angles.

\begin{figure}[tb]
 \vspace{0.5em}
    \centering
    \begin{subfigure}[t]{0.4004\columnwidth}
        \centering
        \includegraphics[width=\columnwidth]{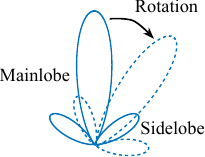}
        \caption{}
        \label{fig:mainlobe}
    \end{subfigure}
    \hspace{0.02\textwidth} 
    \begin{subfigure}[t]{0.2950\columnwidth}
        \centering
        \includegraphics[width=\columnwidth]{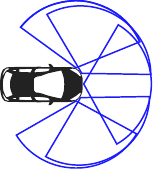}
        \caption{}
        \label{fig:polar}
    \end{subfigure} 
    \hspace{0.02\textwidth} 
    \begin{subfigure}[t]{0.6675\columnwidth}
    	\vspace{2mm}
        \centering
        \includegraphics[width=\columnwidth]{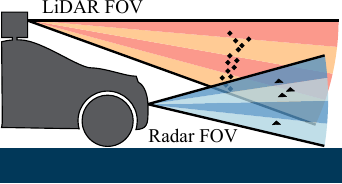}
        \caption{}
        \label{fig:side}
    \end{subfigure}
    \caption{(a) Rotating points disrupts the radiation pattern. (b) Points from different radar sensors adhere to their respective local polar distributions. (c) Mixing points based on elevation angles may be ineffective due to the low angular resolution of radar sensors. }
    \label{fig:problem_explaination}
    \vspace{-1.6em}
\end{figure}

 However, all these approaches have been tested exclusively on LiDAR point cloud data. To apply LiDAR-based MSDA to radar, several challenges must be addressed, as shown in Fig.~\ref{fig:problem_explaination}. In the following paragraphs, we discuss the feasibility of adapting these methods to radar point clouds.
\begin{figure*}[t]
    \vspace{0.5em}
	\centering
	\fbox{%
		\begin{minipage}{0.9\textwidth}
			\hspace{1.0mm}
			\pointexcluded \hspace{0.1mm} Excluded points
			\hfill
			\pointsource \hspace{0.1mm} $s^A$ points
			\hfill
			\pointtarget \hspace{0.1mm} $s^B$ points
			\hfill
			\pointrotated \hspace{0.1mm} Rotated points
			\hfill
			\bboxline \hspace{0.1mm} BBoxes \hspace{0.5mm}
		\end{minipage}%
	}\vspace{2mm}
	\begin{subfigure}[b]{\textwidth}
		\centering
		\includegraphics[width=0.315\textwidth]{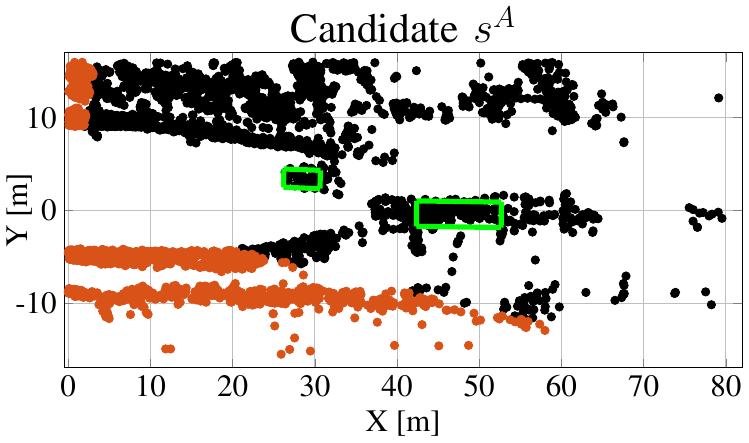}
		\hfill
		\includegraphics[width=0.315\textwidth]{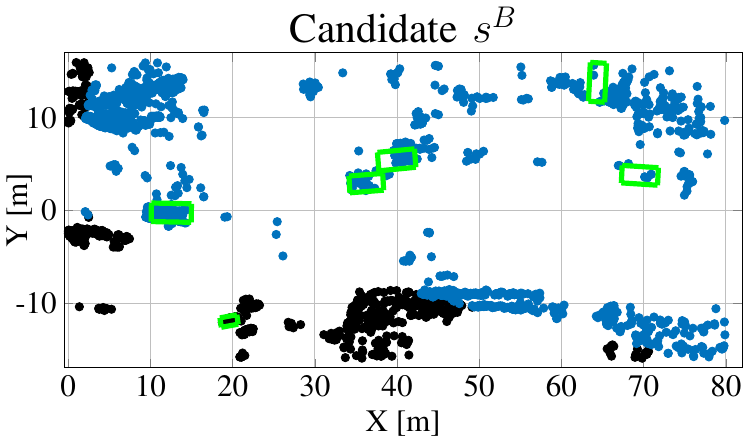}
		\hfill
		\includegraphics[width=0.315\textwidth]{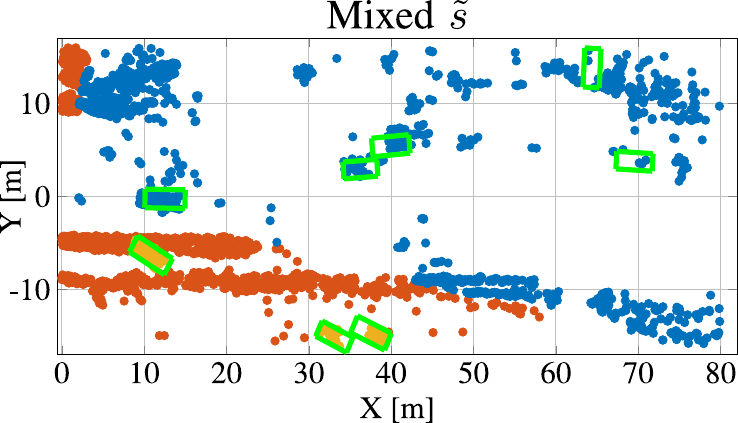}
        \vspace{-1mm}
		\caption{PolarMix (Bird's-Eye View)}
		\label{fig:polar_mix}
	\end{subfigure} 
    
	\begin{subfigure}[b]{\textwidth}
		\centering
		\includegraphics[width=0.3125\textwidth]{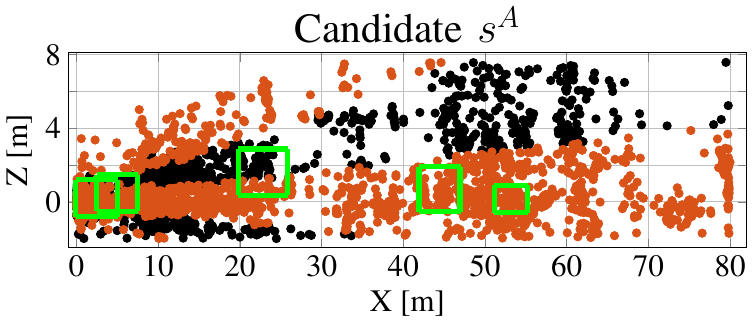}
		\hspace{3.35mm}
		\includegraphics[width=0.3125\textwidth]{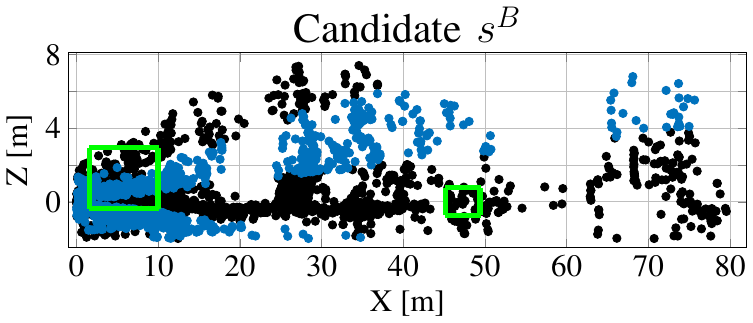}
		\hspace{3.35mm}
		\includegraphics[width=0.3125\textwidth]{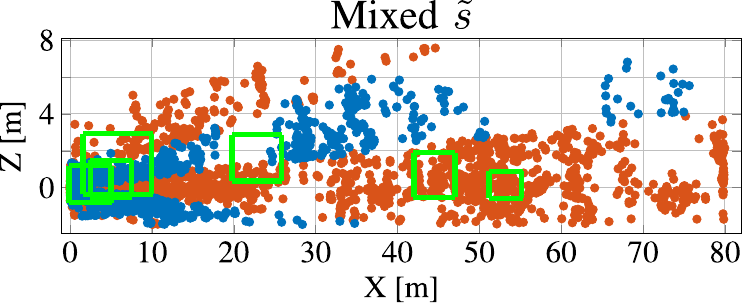}\vspace{-1mm}
		\caption{LaserMix (Side View)}
		\label{fig:laser_mix}
	\end{subfigure}
    \caption{Examples of LiDAR-based MSDA methods applied to radar point clouds in Bosch Street dataset. The point clouds are aggregated from five sensors. In PolarMix, a 90$^\circ$ region along the azimuth axis is extracted and swapped between $s^A$ and $s^B$, while object-level points from $s^B$ are copied and rotated into the new sample $\tilde{s}$. In LaserMix, the point clouds are divided into six regions, and points from non-adjacent areas are combined from $s^A$ and $s^B$.}

	\label{fig:mixup_techniques}
    \vspace{-1.5em}
\end{figure*}

\begin{itemize}[label=$\bullet$,wide = 0pt]
\setlength{\itemindent}{0.5em} \setlength{\itemsep}{1.0pt} \setlength{\parskip}{0.5pt}
 \item \textbf{MixUp~\cite{zhang2017mixup}} was adapted for LiDAR point clouds in~\cite{wang2023ssda3d}.  Binary masks $K$ and $Q$  are employed to select the points such that $\frac{|K=1|}{|K|}=\lambda$ and $\frac{|Q=1|}{|Q|}=1-\lambda$, where $\lambda$ is sampled from a given distribution $\beta$. The output is computed as $\tilde{s}=(K\odot s^A)\oplus(Q\odot s^B)$ and the new label $\tilde{y}=y^A\oplus y^B$. Here, $\odot$  and $\oplus$ represent masking operation and concatenation, respectively. A common choice of $\beta$ distribution is $\beta(2,2)$, which concentrates most values around 0.5. \textbf{\textit{However, radar point clouds can be extremely sparse in certain regions, meaning that applying a single mix ratio uniformly across all points may be suboptimal and limit diversity.}}

\item \textbf{PolarMix~\cite{xiao2022polarmix}} randomly divides the field of view into two azimuth regions and swaps all points within the selected angular interval. As shown in Fig.~\ref{fig:polar_mix}, a random 90$^\circ$ region is chosen for this exchange, with all points and corresponding labels being swapped. Additionally, points within bounding boxes in $s^B$ are copied and rotated around the origin by a randomly selected angle within a predefined range into $\tilde{s}$ to enhance object diversity.
Unlike LiDAR sensors, which offer uniform 360$^\circ$ detection, radar point clouds encounter irregularities in their angular distribution due to the radiation pattern, as shown in Fig.~\ref{fig:mainlobe}. This leads to significant variations in detection capability across the main lobe and side lobes, especially along the azimuth axis. \textbf{\textit{Accordingly, substantial rotation operations such as PolarMix will disrupt the inherent angular distribution of the point cloud.}}

\item \textbf{LaserMix~\cite{kong2023lasermix}} exploits the inclination angle for partitioning by leveraging the distinctive pattern distribution of laser beams.
As explained in Fig.~\ref{fig:laser_mix}, the points clouds are divided into $N$ equally spaced segments based on the elevation angular interval. An alternating selection of point segments from  $s^A, s^B$ is then utilized to construct $\tilde{s}$.
However, multi-radar setups are frequently adopted to enhance radar perception, for instance in the Bosch Street dataset~\cite{armanious2024bosch}, which employs five front-facing radars. As Fig.~\ref{fig:polar} shows, these multiple sensors emit beams from different coordinate systems, so the spatial priors established by laser partitions mentioned in LaserMix~\cite{kong2023lasermix} no longer hold. Additionally, radar sensors typically have a lower resolution in the elevation angle and a narrower vertical field of view compared to LiDAR~\cite{han2023radar} (Fig.~\ref{fig:side}). \textbf{\textit{Thus, mixing based on elevation angle or polar coordinates may not be applicable to radar sensors due to their limited elevation performance and the effects of multi-sensor aggregation.}}

\item \textbf{PillarMix~\cite{wu2024pillarmix}} directly switches the points and labels within selected pillars to generate mixed samples 
Compared to LiDAR point clouds, radar point clouds are significantly sparser. Replacing portions of these sparse objects with irrelevant or empty pillars may remove critical information needed for accurate predictions. Consequently, training on such augmented data can degrade performance, making PillarMix less effective than MixUp for these objects. \textbf{\textit{Hence, directly switching entire pillars may be suboptimal for radar point clouds, especially when dealing with sparse objects.}}
\end{itemize}

As discussed above, the distinct point distribution patterns in radar point clouds, combined with their inherent sparsity and the use of multi-radar sensor fusion, differ significantly from 360$^\circ$ LiDAR data. Therefore, aforementioned methods developed for LiDAR data often fall short when applied to radar. Later results will reveal that some of these approaches not only underperform compared to general mixed-sample techniques like MixUp~\cite{zhang2017mixup} but can even negatively impacting perception performance. 
\section{Approach}
\begin{figure*}[t]
  \centering
  \vspace{0.5em}
  \subfloat[Overview]{%
    \includegraphics[width=0.62\textwidth]{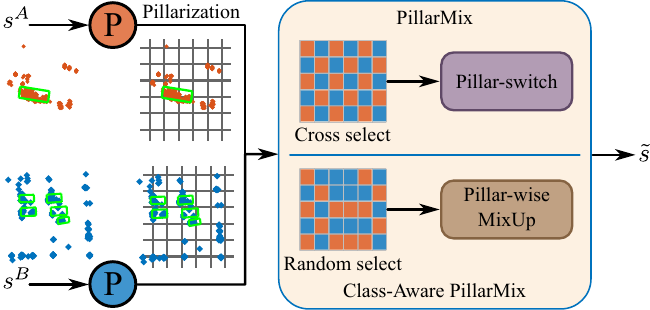}
    \label{fig:overview}
  }
  \hfill
  \subfloat[Pillar-wise MixUp]{%
    \includegraphics[width=0.31\textwidth]{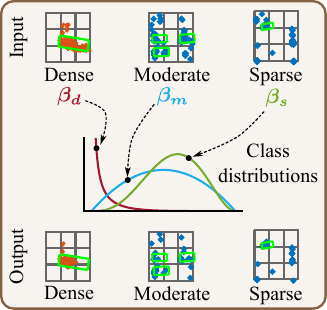}
    \label{fig:pillarwise}
  }
  \caption{Overall framework comparison of PillarMix and our proposed CAPMix. Both methods share the pillarization step, but PillarMix simply switches pillars between two samples $s^A, s^B$.  In contrast, CAPMix randomly selects pillars and then performs class-aware MixUp using different $\beta$ distributions $\textcolor{sparse}{\beta_{s}},\textcolor{moderate}{\beta_{m}},\textcolor{dense}{\beta_{d}}$ based on the classes in the pillars.}
  \label{fig: methods}
\vspace{-1.5em}
\end{figure*}

\subsection{Overview}
Inspired by PillarMix~\cite{wu2024pillarmix}, we propose CAPMix, a class-aware variant that accounts for the sparsity of radar point clouds. The overall methods of both PillarMix and CAPMix are illustrated in Fig.~\ref{fig: methods}. Like PillarMix, CAPMix begins by pillarizing two point clouds $s^A, s^B$ using a specified pillar size $\phi$ (covering an area of $\phi\times\phi$ in Bird's-Eye View). However, instead of using non-adjacent cross-selection as in PillarMix, all the pillars $P$ are randomly divided into $[P_{c},P_{u}]$ based on a ratio $\alpha$. Next, MixUp is applied to each pillar, with the mix ratio independently sampled from a predefined $\beta$ distribution. For $P_{c}$, the mix ratio is class-aware, with pillars assigned a mix ratio from
$\textcolor{sparse}{\beta_{s}},\,\textcolor{moderate}{\beta_{m}},\,\textrm{or}\,\,\textcolor{dense}{\beta_{d}}$ based on their object density (sparse, moderate, or dense).
To enhance diversity while preserving original information, $\textcolor{moderate}{\beta_{m}}$ is also applied to all pillars in $P_{u}$ without considering class labels.
As a data augmentation technique, CAPMix is employed exclusively during training, while the inference stage remains identical to the standard process.
\subsection{Pillar-wise MixUp \label{sec: pillar-wise MixUp}} 
\begin{figure}[tbp]
    \centering
    \vspace{0.5em}
    \includegraphics[width=\columnwidth]{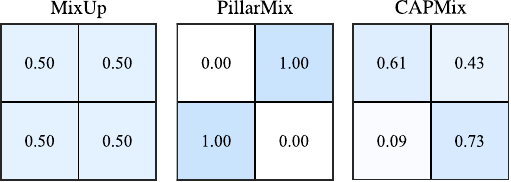}
    \caption{Comparison of the mixing effect, where grid cells represent pillars and numbers indicate the mix ratio $\lambda$.}
    \label{fig:mixing_effect}
    \vspace{-1.2em}
\end{figure}

To mitigate the loss of critical information, as discussed in Sec.\ref{sec: motivation}, we apply MixUp within individual pillars instead of swapping all points as done in PillarMix. As illustrated in Fig.~\ref{fig:mixing_effect}, standard MixUp employs a uniform mix ratio $\lambda$ across all points. Conversely, we apply MixUp within each pillar using an independently sampled mix ratio from a $\beta$ distribution. For instance, $\lambda$ can be 0.61 in one pillar but 0.09 in another, resulting in varied mixing patterns, especially among adjacent pillars. Consequently, the samples become significantly more diverse compared to using a uniform mix ratio across all points as MixUp. Compared with PillarMix, our pillar-wise MixUp retains more original information, making it more suitable for sparse objects than directly swapping all the points within the pillars.
\subsection{Class-aware \texorpdfstring{$\beta$}{beta} distributions in \texorpdfstring{$P_c$}{Pc} \label{sec: capmix}}
\begin{figure*}[t]
        \vspace{0.5em}
	\centering
	\fbox{%
		\begin{minipage}{0.9\textwidth}
			\hspace{1.0mm}
			\pointexcluded \hspace{0.1mm} Excluded points
			\hfill
			\pointsource \hspace{0.1mm} $s^A$ points
			\hfill
			\pointtarget \hspace{0.1mm} $s^B$ points
			\hfill
			\bboxline \hspace{0.1mm} BBoxes \hspace{0.5mm}
		\end{minipage}%
	}\vspace{2mm}
	\begin{subfigure}[b]{\textwidth}
		\centering
		\includegraphics[width=0.31\textwidth]{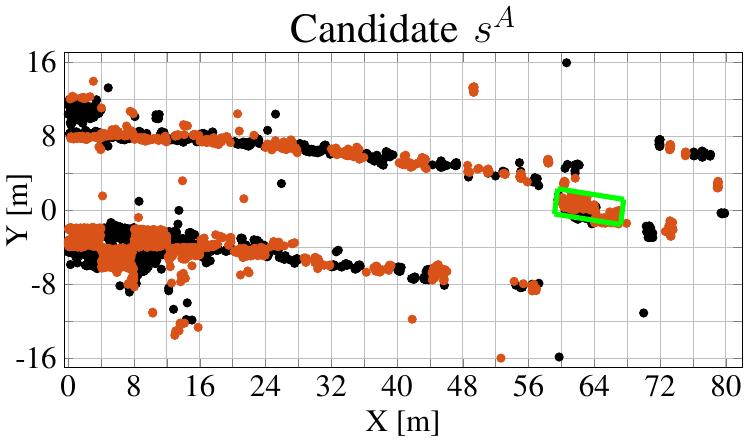}
		\hfill 
		\includegraphics[width=0.31\textwidth]{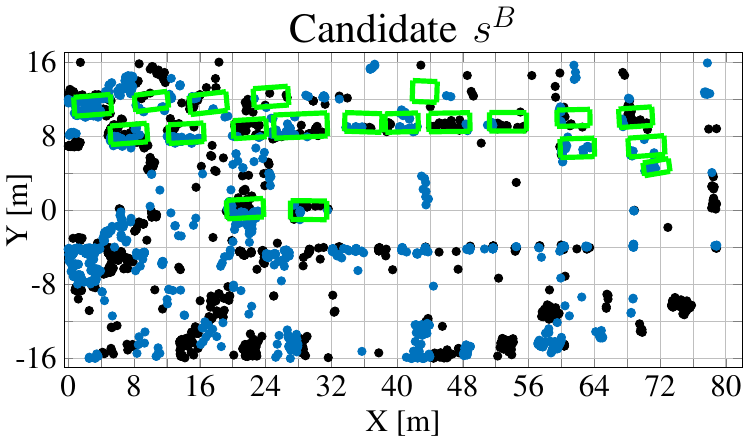}
		\hfill 
		\includegraphics[width=0.31\textwidth]{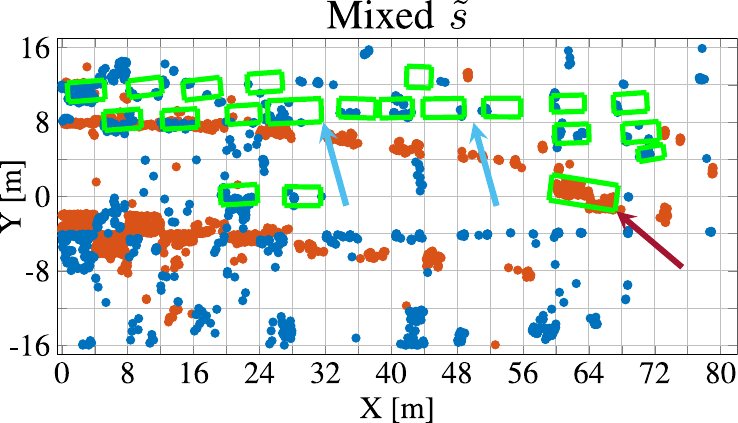}\vspace{-1mm}
		\caption{PillarMix (Bird's-Eye View)}
		\label{fig:pillar_mix}
	\end{subfigure}

	\begin{subfigure}[b]{\textwidth}
		\centering 
		\includegraphics[width=0.31\textwidth]{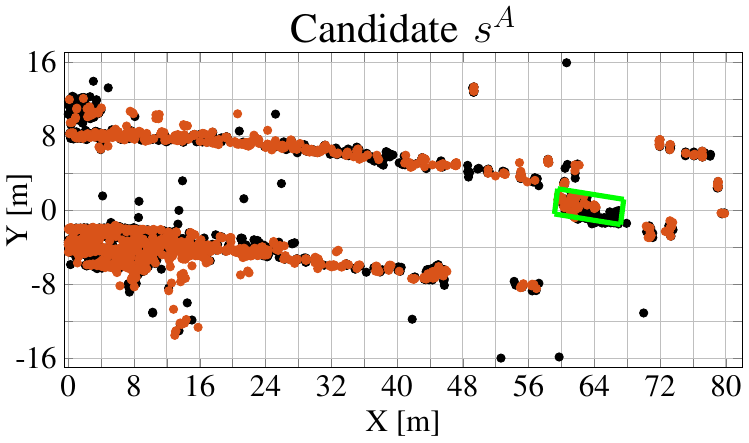}
		\hfill 
		\includegraphics[width=0.31\textwidth]{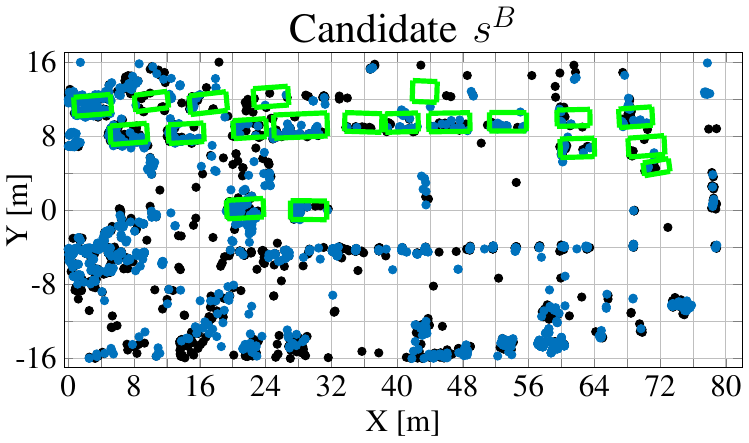}
		\hfill 
		\includegraphics[width=0.31\textwidth]{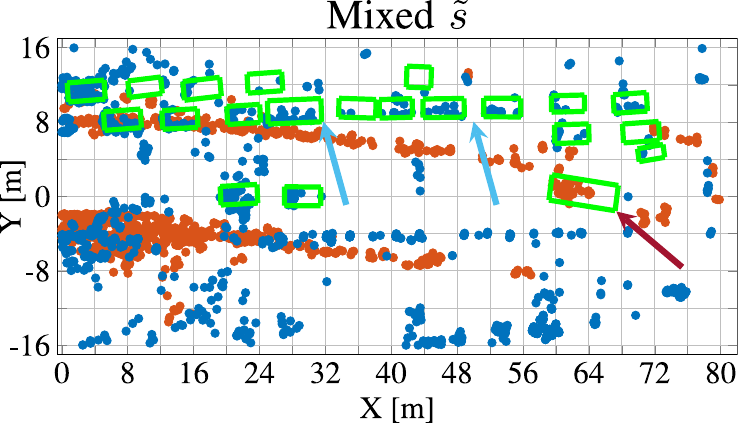}\vspace{-1mm}
		\caption{CAPMix (Bird's-Eye View)}
		\label{fig:cap_mix}
	\end{subfigure}
	\caption{Examples of PillarMix vs. CAPMix for radar point clouds in the Bosch Street dataset.
    In this scenario, the pillar size $\phi$ is set to 4 meters. For car objects, PillarMix’s approach of swapping entire pillars leads to sparsely distributed points, distorting the overall geometry (upper bounding boxes marked by \hspace{0.5mm}\protect\arrowUpModerate\hspace{1mm}). In contrast, CAPMix applies a pillar-wise MixUp with a moderate distribution \textcolor{moderate}{$\beta_m$}, preserving a sufficient number of points for better object integrity. For large vehicles, using a \textcolor{dense}{$\beta_d$} distribution skewed toward 0 produces a similar effect to PillarMix by discarding most points in selected pillars while ensuring that unselected pillars retain enough points (lower bounding boxes marked by \hspace{0.5mm}\protect\arrowUpDense\hspace{1mm}).}
	\label{fig:pillar_vs_cap}
    \vspace{-1.5em}
\end{figure*}

Expanding on the above discussion,  we note a significant variation in point density among different object classes in radar point clouds. For denser objects, there are sufficient points to make robust predictions, enabling us to incorporate more points from another frame during mixing.  In contrast, objects with sparser point clouds carry less information, making them more sensitive to disturbances. Therefore, as explained in Fig.~\ref{fig:pillarwise}, instead of drawing the mix rate from one single $\beta$ distribution, we propose employing different $\beta$ distributions for different classes:
\begin{itemize}
\item Dense objects: For objects like large vehicles, their substantial size and dense point cloud motivate the use of a \textcolor{dense}{$\beta_d$} distribution, which is skewed towards 0. This implies that for pillars in $P_c$, most of the points from the original frame are replaced with points from the other frame. While this approach closely resembles PillarMix, introducing a random mix ratio allows for a broader range of augmentation patterns.
\item Sparse objects: For objects like riders or pedestrians, to mitigate the potential negative effects of mixing, we employ a \textcolor{sparse}{$\beta_s$} distribution that concentrates the majority of its probability density above 0.5. This ensures that a larger proportion of original points is preserved.
\item Moderate objects: For cars and non-object points, we use \textcolor{moderate}{$\beta_m$}, a more symmetric and uniform distribution (same as standard MixUp), to achieve greater diversity in the augmented samples.
\end{itemize}
\begin{algorithm}[!b]
\small
\renewcommand{\algorithmicrequire}{\textbf{Input:}}
\renewcommand{\algorithmicensure}{\textbf{Output:}}
\caption{Class-Aware PillarMix.}
\label{algo}
\begin{algorithmic}[1]
\Require
Points and labels of two radar samples: $\{s^A, \,y^A\}, \{s^B,\, y^B\}$; Pillar Size $\phi$; Pillar select ratio $\alpha$; MSDA probability $\delta$; MixUp $\beta$ distribution $\textcolor{sparse}{\beta_{s}},\,\textcolor{moderate} {\beta_{m}},\,\textcolor{dense}{\beta_{d}}$ for $cls_{moderate}, cls_{sparse},cls_{dense}$.
\Ensure A new radar sample for training: $\{\tilde{s},\, \tilde{y}\}$.
\State $\tilde{s},\, \tilde{y}=s^A,\, y^A$  \qquad \textcolor{gray}{\#
\textit{Initialization}}
\If{rand() $\leq \delta$} \qquad \textcolor{gray}{\# \textit{do MSDA}}
\State $P_A,\,P_B=\textrm{Pillarization}(s^A,\,\phi),\,\textrm{Pillarization}(s^B,\,\phi)$
\State $[P_{c},\,P_{u}]=\textrm{RandomSelect}([P_A,\,P_B],\,\alpha),$
\For{$p\;in\;P_{c}$} \qquad \textcolor{gray}{\# \textit{different mix based on priority}}
\State if $cls_{sparse}\;\textrm{in} \;p_A \,\textrm{or}\;p_B$: $\tilde{p}=MU(p_A,\,p_B,\,\textcolor{sparse}{\beta_{s}})$,  
\State elif $cls_{moderate}\;\textrm{in} \;p_A \,\textrm{or}\;p_B$: $\tilde{p}=MU(p_A,\,p_B,\,\textcolor{moderate}{\beta_{m}})$, 
\State elif $cls_{dense}\;\textrm{in} \;p_A \, \textrm{or}\;p_B$: $\tilde{p}=MU(p_A,\,p_B,\,\textcolor{dense}{\beta_{d}})$, 
\State else: $\tilde{p}=MU(p_A,\,p_B,\,\textcolor{moderate}{\beta_{m}})$
\EndFor
\For{$p\;in\;P_{u}$} \qquad \textcolor{gray}{\# \textit{also mix up unselected pillar}}
\State $\tilde{p}=MU(p_A,\,p_B,\,\textcolor{moderate}{\beta_{m}})$,
\EndFor
\State Update $\tilde{s},\,\tilde{y}$ by concatenating the points from every $\tilde{p}$ along with the labels of both samples
\EndIf
\end{algorithmic}
\end{algorithm}

In cases where a single pillar includes different class objects, operations are determined based on priority levels. Sparse objects receive higher priority since they are more sensitive to augmentation operations, thus requiring greater care to preserve their features. On the other hand, the dense objects hold the lowest priority level, since \textcolor{dense}{$\beta_d$} is close to 0.

 To ensure the model retains sufficient information for accurate predictions, the class-aware mix ratio is applied only to $P_c\,$, rather than all pillars. Applying it universally would result in excessive information loss, particularly for large objects, making them significantly harder for the model to detect. For instance, if $\textcolor{dense}{\beta_{d}}$ were applied to dense objects in $P_u$, every part of the object would receive a $\lambda$ value close to 0, leading to the removal of most points. In contrast, using $\textcolor{moderate}{\beta_{m}}$ for $P_u$ ensures that enough points are preserved, maintaining the integrity of the object and facilitating reliable predictions. Compared to PillarMix~\cite{wu2024pillarmix}, which retains all points in $P_u\,$, applying a moderate mix ratio provides a balanced approach, preserving sufficient information for effective detection while still benefiting from augmentation. The operation can be mathematically expressed as follows:
 \begin{equation}
 \vspace{1mm}
\begin{aligned}
\centering 
    \tilde{p}_{c}&= MU(p_A,\,p_B,\,\beta),\, \beta \in (\textcolor{sparse}{\beta_{s}},\,\textcolor{moderate}{\beta_{m}},\,\textcolor{dense}{\beta_{d}}),\\
    \tilde{p}_{u}&= MU(p_A,\,p_B,\,\textcolor{moderate}{\beta_{m}}),\\
    MU(p_A,\,p_B,\,\beta)&=(K\odot p_A)\oplus(Q\odot p_B),\\
    \tilde{s}&= \{\tilde{p}_c\}\oplus\{\tilde{p}_u\}, \\
    \tilde{y}&= y^A \oplus y^B.
\end{aligned}
\label{eq:capmix} \vspace{1mm}
 \end{equation}
Here, $p_A$ and $p_B$ represent pillars extracted from two input point clouds, $s^A$ and $s^B$. As in the MixUp ($MU$) operation described in Sec.~\ref{sec: motivation}, binary masks $K$ and $Q$ are used to select points for mixing. The mixed points from $P_c$ (the selected pillars) are denoted as $\tilde{p}_{c}$, while $\tilde{p}_{u}$ represents the mixed points from $P_u$ (the unselected pillars). The complete process is outlined in Algorithm~\ref{algo}.

 Our method leverages bounding box labels while operating on radar data pillars, recognizing that radar points can extend beyond the bounding box, particularly for moving objects ~\cite{palmer2023ego}. By applying pillar-wise operations, we ensure that these points, along with non-object points, are included in the mixing process. Moreover, we use object labels rather than sparsity to distinguish pillars, ensuring sparse objects are not overshadowed or erased when sharing a pillar with dense objects. This careful handling of sparse points with high priority is crucial for maintaining data integrity.

Fig.~\ref{fig:pillar_vs_cap} compares the point clouds processed by PillarMix and CAPMix. In PillarMix, some car objects become overly sparse as they distribute across multiple pillars. Whereas, CAPMix preserves more points for accurate prediction. For dense objects, points in $P_c$ are dropped with a mix ratio from $\textcolor{sparse}{\beta_s}$, while a sufficient number of points in $P_u$ are retained with a mix ratio from $\textcolor{moderate}{\beta_m}$.

\section{Experiments}
\begin{table*}[tbp]
\vspace{0.5em}
\centering
\caption{Comparison of $AP_{BEV}$/$AP_{3D}$ of different MSDA methods on Bosch Street dataset. \label{tab: bsd}}
\resizebox{\textwidth}{!}{
\begin{threeparttable}
    \centering
    \Huge
    \begin{tabular}{l|cccc|c|cccc|c}
        \toprule
        \multirow{2}{*}{Methods} & \multicolumn{5}{c|}{5\% Data} & \multicolumn{5}{c}{10\% Data}\\
        \cline{2-6} \cline{7-11}& Car & Large Vehicle & Pedestrian & Ridable Vehicle & mAP& Car & Large Vehicle & Pedestrian & Ridable Vehicle & mAP \\ 
        \midrule
        \rowcolor{lightgreen!75} 
        w/o MSDA   &66.09/56.15&34.24/30.36&11.91/7.66&15.94/12.01&32.09/26.56& 70.07/60.90 &41.61/38.30 &15.81/10.23 & 19.66/15.84& 36.79/31.32 \\ 
        \midrule
        CutMix~\cite{yun2019cutmix}    & 66.05/56.71 & 35.98/32.40 & 11.82/7.70 & 15.83/11.79 & 32.42/27.15& 70.78/61.42 & 44.04/39.99& 15.77/10.31 & 19.53/15.26& 37.53/31.78 \\ 
        MixUp~\cite{zhang2017mixup}     & \underline{66.11/56.81} & 39.01/34.89& 15.46/9.81&\underline{17.81}/13.48 &\underline{34.60/28.73}& 70.73/\underline{61.94} & 46.13/42.32 & \underline{17.11/11.12} & \underline{21.50/16.82} & \underline{38.86}/32.93 \\ 
        LaserMix~\cite{kong2023lasermix}  & 66.05/55.95 & 38.43/33.84 & 14.85/9.22 & 17.73/\underline{13.61} & 34.27/28.16& \underline{70.81}/61.42 & 46.36/42.50 & 16.49/10.63 & 20.63/16.24 & 38.57/32.70\\
        PolarMix~\cite{xiao2022polarmix} & 65.27/56.25& 35.40/31.23 & 13.42/8.83 & 16.84/12.36& 32.73/27.17& 69.97/60.73 & 44.28/40.21 & 16.03/10.18 & 20.28/16.11& 37.64/31.81\\
        PillarMix~\cite{wu2024pillarmix}& 64.87/55.73 & \underline{40.22/35.89} & \textbf{15.58}/\underline{9.98} & 15.78/11.47 & 34.11/28.27& 70.23/61.17 & \textbf{48.15/44.54}& 16.70/10.92& 19.90/15.56 & 38.75/\underline{33.05} \\
        \midrule
        \textbf{CAPMix (Ours)}&\textbf{66.47/57.21} & \textbf{41.29/37.57} & \underline{15.55}/\textbf{10.12}& \textbf{18.53/14.27}& \textbf{35.46/29.79}& \textbf{71.01/62.03}&\underline{47.72/43.86}&\textbf{17.22/11.26}&\textbf{ 21.64/16.92}&\textbf{39.40/33.52}\\
       \textcolor{olive}{ \makecell[c]{$\triangle \uparrow$}} &\textcolor{olive}{+0.38/+1.06}& \textcolor{olive}{+7.05/+7.21}& \textcolor{olive}{+3.64/+2.46}& \textcolor{olive}{+2.59/+2.26}& \textcolor{olive}{+3.37/+3.15}&\textcolor{olive}{+0.94/+1.13}& \textcolor{olive}{+6.11/+5.56}& \textcolor{olive}{+1.41/+1.03}&\textcolor{olive}{+1.98/+1.08}&\textcolor{olive}{+2.61/+2.20}\\
        \midrule
        \multirow{2}{*}{Methods} & \multicolumn{5}{c|}{25\% Data} & \multicolumn{5}{c}{50\% Data}\\
        \cline{2-6} \cline{7-11}& Car & Large Vehicle & Pedestrian & Ridable Vehicle & mAP& Car & Large Vehicle & Pedestrian & Ridable Vehicle & mAP \\ 
        \midrule
        \rowcolor{lightgreen!75} 
        w/o MSDA   & 74.33/65.86& 48.60/44.40&17.74/11.94& 22.06/17.72&40.67/34.98&75.86/67.82&52.84/49.00& 19.26/13.56& 25.44/21.37 & 43.35/37.93 \\ 
        \midrule
        CutMix~\cite{yun2019cutmix}    & 74.39/66.30& 49.20/44.95 &17.75/11.96&23.11/18.55 &41.11/35.44&76.47/69.11&53.22/49.21&20.38/14.57& 26.44/22.17 & 44.13/38.76\\ 
        MixUp~\cite{zhang2017mixup}     & \underline{74.57/66.44} & 51.51/47.59& \textbf{19.27/13.25}&\textbf{25.07/20.47}&\underline{42.60/36.93}& \underline{76.49/69.37} & \underline{55.80/52.05}& \textbf{21.14/15.27} & \underline{28.28/23.70}& \underline{45.43/40.10}\\ 
        LaserMix~\cite{kong2023lasermix}  & 74.53/65.48& 50.52/46.59 &18.69/12.22&24.62/19.73&42.08/35.93&76.25/67.77&54.72/50.75&20.46/13.92&28.24/23.08& 44.92/38.88\\
        PolarMix~\cite{xiao2022polarmix} & 73.97/66.03& 47.77/44.38 &18.10/12.28& 23.47/19.16 &40.82/35.46&74.89/67.67&52.80/48.45&20.10/14.40& 27.34/22.62& 43.78/39.29\\
        PillarMix~\cite{wu2024pillarmix}& 74.23/66.00& \underline{52.38/48.47} &18.89/\underline{13.15}&23.63/18.83 &42.28/36.61&76.15/68.89&55.50/51.68&20.57/14.79& 27.15/22.26 &44.84/39.30\\
        \midrule
        \textbf{CAPMix (Ours)}& \textbf{74.64/66.45}&\textbf{52.49/48.87}&\underline{19.12}/13.13& \underline{25.03/20.21} &\textbf{42.82/37.17}&\textbf{76.62/69.54}&\textbf{56.49/53.16}&\underline{ 21.11/15.21}&\textbf{28.81/24.22}&\textbf{45.73/40.43}\\
       \textcolor{olive}{ \makecell[c]{$\triangle \uparrow$}} &\textcolor{olive}{+0.31/+0.59}& \textcolor{olive}{+3.89/+4.47}& \textcolor{olive}{+1.38/+1.19}& \textcolor{olive}{+2.97/+2.49}& \textcolor{olive}{+2.15/+2.19}&\textcolor{olive}{+0.76/+1.72}& \textcolor{olive}{+3.65/+4.16}& \textcolor{olive}{+1.85/+1.65}&\textcolor{olive}{+3.37/+2.85}&\textcolor{olive}{+2.38/+2.50}\\
        \midrule
        \rowcolor{lightgray!50} 
        100\% data&77.27/70.06&57.26/53.46&20.14/14.62&28.26/24.00&45.73/40.54& 77.27/70.06&57.26/53.46&20.14/14.62&28.26/24.00&45.73/40.54\\ 
        \bottomrule
    \end{tabular}
    \begin{tablenotes}
    \item The \textbf{best} and \underline{second best} scores are highlighted in \textbf{bold} and \underline{underline}. The symbol \textcolor{olive}{$\triangle \uparrow$} indicates the increase compared to the baseline without MSDA.
    \end{tablenotes}
\vspace{-2em}
\end{threeparttable}
}
\vspace{2mm}
\end{table*}
\subsection{Datasets}
For a comprehensive analysis, we examine our methods on two datasets:
\begin{itemize}[label=$\bullet$,wide = 0pt]
\setlength{\itemindent}{0.5em} \setlength{\itemsep}{1.0pt} \setlength{\parskip}{0.5pt}
\item \textbf{Bosch Street~\cite{armanious2024bosch}} is a large-scale, multi-modal dataset featuring imaging radar. It comprises 1.3 million frames (36.5 hours) collected across nine cities under three weather conditions. We extract a subset by subsampling at 1-second intervals, resulting in 81k frames for training and 24k for validation, which serve as the $100\%$ data baseline in subsequent experiments. We conduct experiments on datasets of different scales, where the subsets are generated by randomly selecting entire sequences rather than individual frames, reflecting a practical data collection workflow. All results are reported on the validation set. To obtain higher-quality point clouds, we apply a five-frame aggregation strategy.
For more details, we refer readers to the original paper~\cite{armanious2024bosch}.
\item \textbf{K-Radar~\cite{paek2022k}} contains over 35k frames of around one hour recorded in two cities under seven weather conditions. As noted previously, in the original dataset split, frames in the training and test sets were sampled from the same sequences~\cite{fent2024dpft,zhang2024exploring}. To address this, we restructured the dataset by sequences, assigning 44 sequences (27k frames) to the training set and the remaining 24 sequences (7.9k frames) to the test set. The new split ensures that all weather conditions are present in both the training and test sets. Since the number of sequences is limited, the subsets for experiments are randomly sampled by frames. For runtime efficiency, we generate the sparse point cloud using constant false alarm rate (CFAR) thresholding on the raw 4D radar cube, rather than deriving points only from the top 10\% of power values as in the original paper, which aligns with standard radar signal processing practices~\cite{han2023radar}. 
\end{itemize}
\subsection{Implementation Details}
For evaluation, we adopt CenterPoint~\cite{yin2021center} as our 3D detector. 
The detection range for the $x$, $y$, and $z$ axes is defined as $[0, 80] \times [-16, 16] \times [-2, 7.6]$ meters for both datasets. Following K-Radar~\cite{paek2022k} settings,  we utilize the AdamW optimizer along with a Cosine Annealing learning rate scheduler and evaluate performance using $AP_{BEV}/AP_{3D}$ with an IoU threshold of 0.3 for all classes. Unlike previous setups, we report $AP|_{R_{40}}$ instead of $AP|_{R_{11}}$ as it offers a more representative evaluation~\cite{simonelli2019disentangling}. 

Moving on to the hyperparameter settings of CAPMix,  we set the pillar size $\phi=2$ meters (i.e., covering an area of 4 square meters in BEV) for all pillar-based MSDA. As for the choice of class-aware $\beta$ distribution, we use $\textcolor{moderate}{\beta_m}=\beta(2,2)$, $\textcolor{sparse}{\beta_s}=\beta(4,3)$, $\textcolor{dense}{\beta_d}=\beta(0.1,5)$ based on empirical experiments (see Sec.~\ref{sec: beta}). For the Bosch Street dataset, four classes are partitioned as follows: Car $\rightarrow \textcolor{moderate}{\beta_m}$, Large Vehicle $\rightarrow \textcolor{dense}{\beta_d}$, Pedestrian, Ridable Vehicle  $\rightarrow \textcolor{sparse}{\beta_s}$. For K-Radar, only two classes are considered: Sedan $\rightarrow \textcolor{moderate}{\beta_m}$, Bus/Truck $\rightarrow \textcolor{dense}{\beta_d}$. To ensure a fair comparison, we follow the procedure outlined in PillarMix~\cite{wu2024pillarmix} by applying random rotations and scaling to all methods, including baseline without MSDA. All methods are implemented using the MMDet3D framework~\cite{mmdet3d2020}. The reported results represent the mean of three runs with different random seeds on NVIDIA A100 GPU.
\section{Results and Discussion}
\subsection{Bosch Street Dataset}
Table~\ref{tab: bsd} presents a comparison of different MSDA methods applied to 5\%, 10\%, 25\%, and 50\% of the training data. To understand the effectiveness of these methods, we first examine their impact across different object classes. Although MSDA techniques generally improve performance, the degree of enhancement varies. In particular, the car class shows relatively minor gains, likely due to its extensive number of annotations. Even with just 5\% of the data, there are still 22,334 annotated cars, compared to only 1,611 large vehicles, 6,495 pedestrians, and 1,816 ridable vehicles. Consequently, an ill-suited MSDA method can actually reduce performance, as seen in 5\% dataset results, where LaserMix~\cite{kong2023lasermix} and PillarMix~\cite{wu2024pillarmix} drop performance by 0.82 and 1.22 on $AP_{BEV}$, respectively.

Conversely, MSDA methods often provide greater benefits for other classes with fewer annotations. For instance, large vehicles show particularly strong improvements, and PillarMix performs exceptionally well in this category, likely due to its ability to handle truncated objects using pillars. However, for smaller objects such as pedestrians and ridable vehicles, PillarMix’s truncation effect is less apparent. In these scenarios, MixUp, a more general MSDA technique, tends to outperform LiDAR-specific methods. Specifically, MixUp achieves superior results over LiDAR-based approaches in pedestrian and ridable vehicle tasks when using 10\%, 25\%, and 50\% of the data. This finding highlights the potential limitations of LiDAR-oriented MSDA strategies when applied to radar point clouds.

Meanwhile, our CAPMix demonstrates competitive performance across all four classes. For the car class, which has the highest number of annotations, CAPMix consistently improves performance, particularly in $AP_{3D}$. Specifically, it achieves improvements of 1.06, 1.13, 0.59, and 1.72 when using 5\%, 10\%, 25\%, and 50\% of the data, respectively. For large vehicles, CAPMix surpasses the without MSDA baseline by 7.05 in $AP_{BEV}$ and 7.21 in $AP_{3D}$, with just 5\% of the data. This gain stems from its class-aware $\beta$ distribution, which mimics the truncation effect of PillarMix without compromising other classes. Turning to small objects (pedestrian and ridable vehicle), CAPMix also delivers outstanding results, achieving the best performance with only 10\% of the data and maintaining top results for ridable vehicles while ranking second for pedestrians at 50\%.

By introducing a class-aware $\beta$ distribution, CAPMix brings together the benefits of MixUp and PillarMix, delivering superior results across all classes. It achieves the highest mAP across four data scales. Notably, with just 25\% of the data, CAPMix matches the mAP of 50\% data without MSDA, and using 50\% of the data exceeds the full dataset baseline in $\textrm{mAP}_{BEV}$. These findings underscore CAPMix’s potential in significantly reducing data collection costs.
\subsection{K-Radar}
We extend our study to K-Radar, which focuses on two vehicle classes (sedan and bus/truck) as considered in the original paper~\cite{paek2022k}. Since K-Radar excludes categories like pedestrian and ridable vehicle, we apply only the $\textcolor{dense}{\beta_d}$ and $\textcolor{moderate}{\beta_m}$ components of CAPMix, leaving its advantage for sparse classes untested. Furthermore, K-Radar’s adverse weather conditions introduce additional challenges beyond those found in Bosch Street data. 

As shown in Table~\ref{tab:kradar}, MSDA methods yield notable gains on K-Radar, mirroring trends observed on Bosch Street data. Likewise, LiDAR-based approaches such as LaserMix and PolarMix lag behind naive MixUp, with PolarMix even falling below the without MSDA baseline on dense classes. This outcome reaffirms that LiDAR-centric MSDA methods may not transfer effectively to radar data. On the other hand, PillarMix remains robust for larger vehicles, and CAPMix consistently excels across both sedan and bus/truck. Specifically, CAPMix achieves the highest mAP, surpassing the without MSDA baseline by 5.72 and 6.70 on 
$AP_{BEV}$ and $AP_{3D}$, respectively, and outperforming the second-best method by 0.35 and 0.06.

Furthermore, using only 10\% of the data with CAPMix can exceed the performance of a model trained on 100\%, highlighting the severe risk of overfitting given K-Radar’s limited sequences. Note that the 10\% subset used is sampled by frames rather than entire sequences, which helps narrow the performance gap compared to Bosch Street data.
\begin{table}[btp]
\vspace{0.5em}
\centering
\resizebox{0.47\textwidth}{!}{
\begin{threeparttable}
    \centering
    \caption{Comparison of $AP_{BEV}$/$AP_{3D}$ of different MSDA methods on K-Radar.}
    \begin{tabular}{l|cc|c}
        \toprule
        \multirow{2}{*}{Methods} & \multicolumn{3}{c}{10\% Data}\\  
        \cline{2-4} & Sedan & Bus/Truck& mAP \\
        \midrule
        \rowcolor{lightgreen!75} 
        w/o MSDA  & 44.67/32.95& 27.41/19.12 & 35.86/26.03\\
        \midrule
        CutMix~\cite{yun2019cutmix}    & 47.60/37.53 & 24.83/15.86  & 36.21/26.69\\ 
        MixUp~\cite{zhang2017mixup}  & \underline{49.28}/\textbf{38.30} & 30.65/23.86 & 39.97/31.09  \\ 
        LaserMix~\cite{kong2023lasermix} & 46.59/36.52& 31.99/25.87 & 39.29/30.35 \\
        PolarMix~\cite{xiao2022polarmix} & 47.02/35.62 & 24.41/14.67  & 35.71/25.14 \\
        PillarMix~\cite{wu2024pillarmix}& 47.89/36.96 & \textbf{34.55/28.38}& \underline{41.22/32.67}\\
        \midrule
        \textbf{CAPMix (Ours)}& \textbf{49.70}/\underline{38.27} & \underline{33.46/26.55}& \textbf{41.58/32.73}\\
        \textcolor{olive}{\makecell[c]{$\triangle \uparrow$}}& \textcolor{olive}{+5.03/+5.32} & \textcolor{olive}{+6.05/+7.43} & \textcolor{olive}{+5.72/+6.70}\\
        \midrule
         \rowcolor{lightgray!50} 
        100\% data&50.45/39.84&25.35/16.95&37.90/28.39\\
        \bottomrule
    \end{tabular}
    \label{tab:kradar}
    \begin{tablenotes}
    \item The \textbf{best} and \underline{second best} scores are highlighted in \textbf{bold} and \underline{underline}. The symbol \textcolor{olive}{$\triangle \uparrow$} indicates the increase compared to the baseline without MSDA.
    \end{tablenotes}
\vspace{-2em}
\end{threeparttable}
}
\end{table}
\subsection{Ablation Study \label{sec: beta}}
To verify the design choices of the proposed approach, we conduct ablation studies using 5\% of the training data from the Bosch Street dataset.
\begin{itemize}[label=$\bullet$,wide = 0pt]
\setlength{\itemindent}{0.5em} \setlength{\itemsep}{1.0pt} \setlength{\parskip}{0.5pt}
\item \textbf{Pillar Size Effect:}
In this part, we examine the impact of pillar size on perception performance. PillarMix~\cite{wu2024pillarmix} has shown that pillar size is crucial, since larger sizes require fewer computations but reduce the effectiveness of edge-truncated augmentation. As shown in Table~\ref{tab:pillar size}, using 8-meter pillars lowers performance across all four classes, with large vehicles experiencing about a 1.4\% drop. This drop can be attributed to larger objects spanning multiple pillars, reducing the total number of distinct pillars and limiting augmentation diversity. It is important to note that, unlike PillarMix (which directly swaps pillars between two frames), we apply MixUp within each pillar. This approach selects random pillars from the entire sample instead of relying on a cross-selection pattern, resulting in more diverse augmented point clouds and enhancing the effectiveness of augmentation.
\begin{table}[bp]
    \vspace{-1.5em}
    \centering
    \caption{Effect of pillar size on $AP_{BEV}$ and $AP_{3D}$.}
    \label{tab:pillar size}
    {\renewcommand{\arraystretch}{1.3}
    \resizebox{\columnwidth}{!}{
      {\Huge
      \begin{tabular}{l|cccc|c}
        \toprule
        Pillar Size & Car & Large Vehicle & Pedestrian & Ridable & mAP\\
        \midrule
        $2\times2\,\textrm{m}^2$ & \textbf{66.47/57.21} & \textbf{41.29/37.57} & 15.55/\textbf{10.12}& \textbf{18.53/14.27} & \textbf{35.46/29.79}\\
        $4\times4\,\textrm{m}^2$ & 66.21/56.86 & 40.59/36.92 & \textbf{15.69}/10.00 & 18.12/13.78 & 35.15/29.39\\
        $8\times8\,\textrm{m}^2$ & 66.20/56.91 & 39.79/36.19 & 15.41/9.88 & 17.98/13.96 & 34.84/29.25\\
        \bottomrule
      \end{tabular}
      }
    }
    }
\end{table}
\item \textbf{Class-Aware MixUp Variants:}
Building on our class-aware approach, we assign different $\beta$ distributions to pillars from different classes to derive mix ratios. In Table~\ref{tab:mix_ratio}, we explore various parameter settings. Notably, even when all pillars share the same $\beta(2,2)$ distribution, performance still surpasses standard MixUp. This improvement stems from each selected pillar receiving its own mix ratio, rather than relying on a single uniform ratio across all pillars.
For sparse $\textcolor{sparse}{\beta_s}$, we found that values should be greater than 0.5 to ensure sufficient retention of the original information. For instance, $\textcolor{sparse}{\beta_s} = \beta(10,1)$ keeps in-pillar augmentation minimal while mixing surrounding pillars with other distributions to enhance data diversity. On the other hand, near-zero distributions, such as $\textcolor{sparse}{\beta_s} = \beta(0.1, 5)$, are not suitable for sparse objects. Meanwhile, $\textcolor{dense}{\beta_d}$, shows better performance with values closer to zero, as this class-specific ratio is applied only within randomly selected pillars. Since large objects often span multiple pillars, only a fraction of them (determined by ratio $\alpha$) are assigned $\textcolor{dense}{\beta_d}$, while the remaining pillars use $\textcolor{moderate}{\beta_m}$ to preserve original frame information and prevent excessive truncation.

Given that moderate and environmental pillars account for the majority of the dataset,  $\textcolor{moderate}{\beta_m}=\beta(2,2)$ serves as a reliable choice for these segments. As mentioned in Sec.~\ref{sec: capmix}, our class-aware $\beta$ distribution is applied only to randomly selected pillars $P_c$, while all unselected pillars $P_u$ use $\textcolor{moderate}{\beta_m}$ for mixing. This strategy ensures that the baseline performance of MixUp is preserved regardless of the choice of $\beta$ parameters. Finally, the highlighted section in Table~\ref{tab:kradar} demonstrates that our configuration achieves a well-balanced trade-off across classes and extends effectively to K-Radar.
\begin{table}[tb]
    \centering
    \vspace{0.5em}
    \caption{Influence of Class-Aware MixUp variants on $AP_{BEV}$ and $AP_{3D}$.}
    \label{tab:mix_ratio}
    \renewcommand{\arraystretch}{1.3}%
    \resizebox{\columnwidth}{!}{
    \Huge
    \begin{tabular}{c|c|cccc|c}
    \toprule
    Methods&Visulization&Car&Large Vehicle&Pedestrian&Ridable Vechicle&mAP\\
    \midrule
     \rowcolor{lightgray!50} 
    w/o MSDA&-&66.09/56.15&34.24/30.36&11.91/7.66&15.94/12.01&32.09/26.56\\
    \midrule
     \rowcolor{lightgray!50} 
    MixUp&-&66.11/56.81 & 39.01/34.89& 15.46/9.81&17.81/13.48& 34.60/28.73\\
    \midrule
     \makecell[l]{
         $\beta_s=\beta(2,2)$ \\ 
         $\beta_d=\beta(2,2)$ \\ 
         $\beta_m=\beta(2,2)$
        }& \makecell{\raisebox{-0.1\height}{\resizebox{0.45\linewidth}{!}{\includegraphics{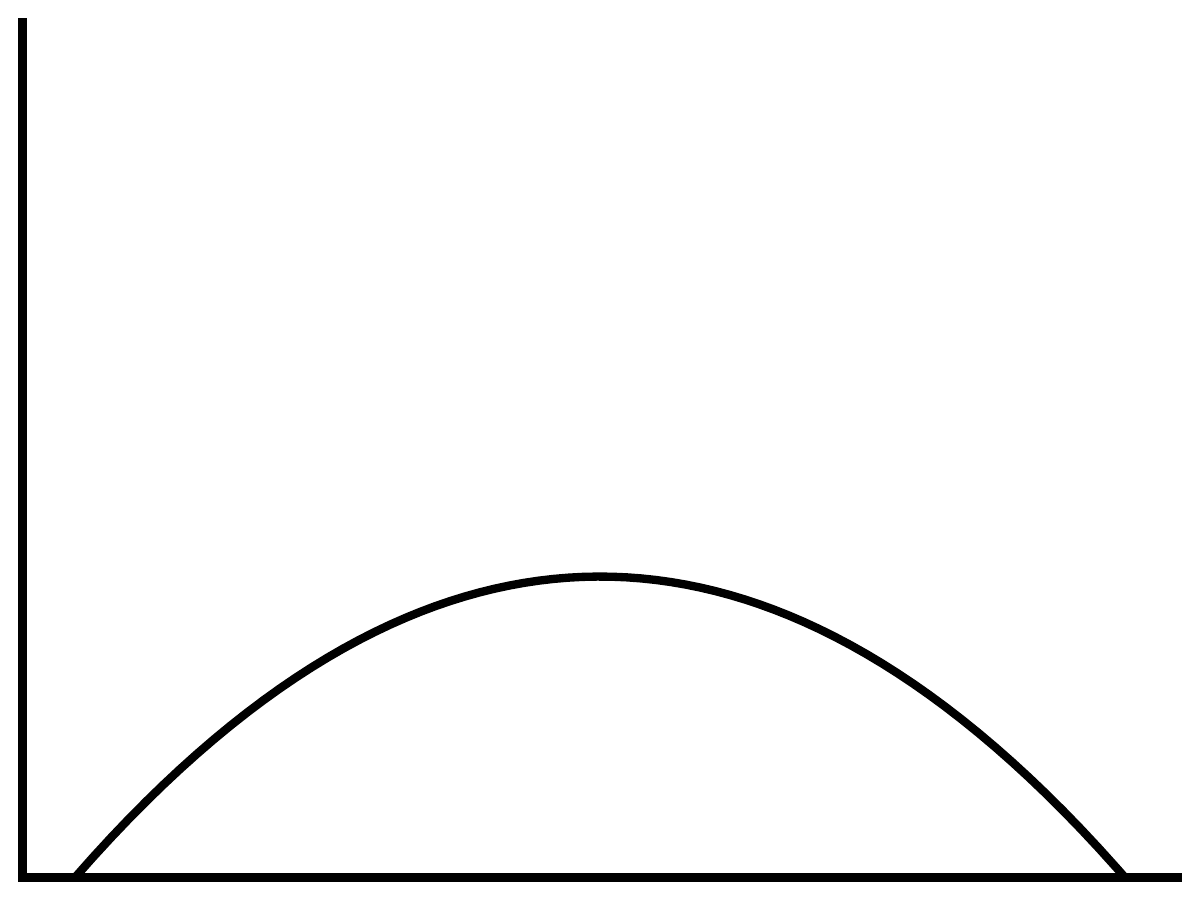}}}}&66.23/56.99&40.34/36.39&15.49/9.97&17.80/13.79&34.95/29.26\\
    \midrule
   \rowcolor{lightgreen!75}  \makecell[l]{
            $\textcolor{sparse}{\beta_s=\beta(4,3)}$\\
         $\textcolor{moderate}{\beta_m=\beta(2,2)}$ \\ 
         $\textcolor{dense}{\beta_d=\beta(0.1,5)}$ 
        } &\makecell{\raisebox{-0.1\height}{\resizebox{0.45\linewidth}{!}{\includegraphics{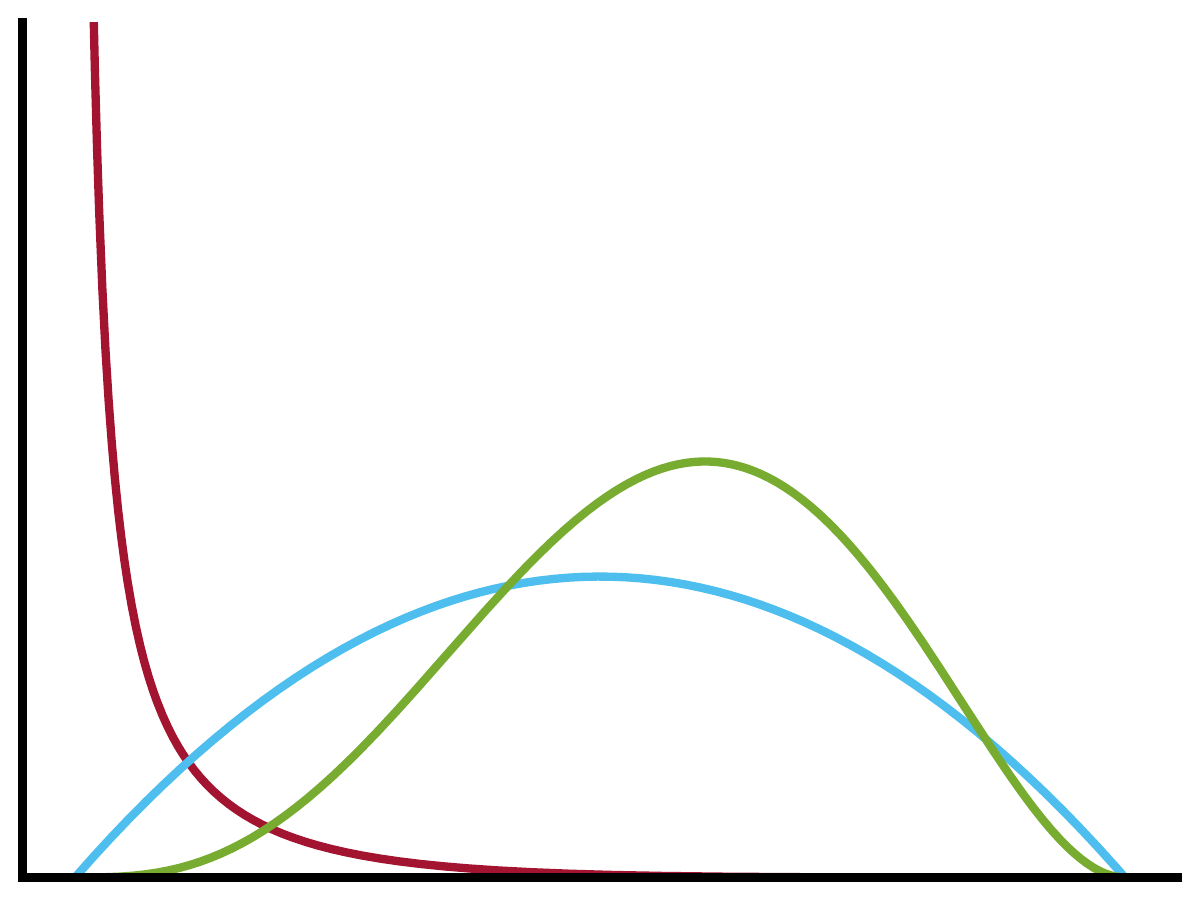}}}}& \textbf{66.47}/57.21 & 41.29/37.57 & 15.55/\textbf{10.12}& \textbf{18.53}/\textbf{14.27}& \textbf{35.46}/\textbf{29.71}\\
    \midrule
    \makecell[l]{
         $\textcolor{sparse}{\beta_s=\beta(5.67,3)}$ \\ 
         $\textcolor{moderate}{\beta_m=\beta(2,2)}$\\
        $\textcolor{dense}{\beta_d=\beta(2,10)}$ 
        } & \makecell{\raisebox{-0.1\height}{\resizebox{0.45\linewidth}{!}{\includegraphics{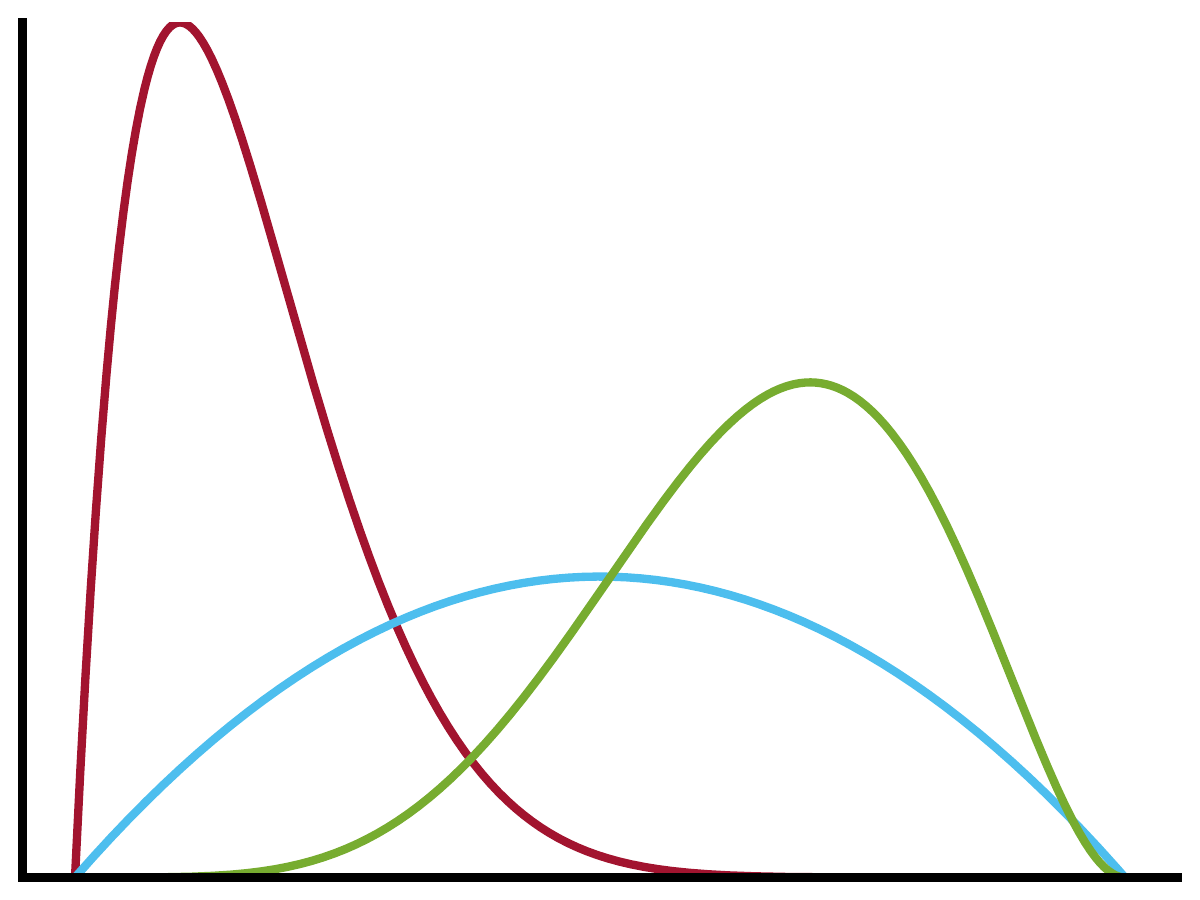}}}}&66.42/\textbf{57.27}&40.55/36.70&15.45/10.05&17.76/13.77&35.04/29.45\\
    \midrule
    \makecell[l]{
         $\textcolor{sparse}{\beta_m=\beta(2,2)}$\\
         $\textcolor{moderate}{\beta_s=\beta(2,2)}$ \\ 
         $\textcolor{dense}{\beta_d=\beta(0.1,5)}$ 
        } & \makecell{\raisebox{-0.1\height}{\resizebox{0.45\linewidth}{!}{\includegraphics{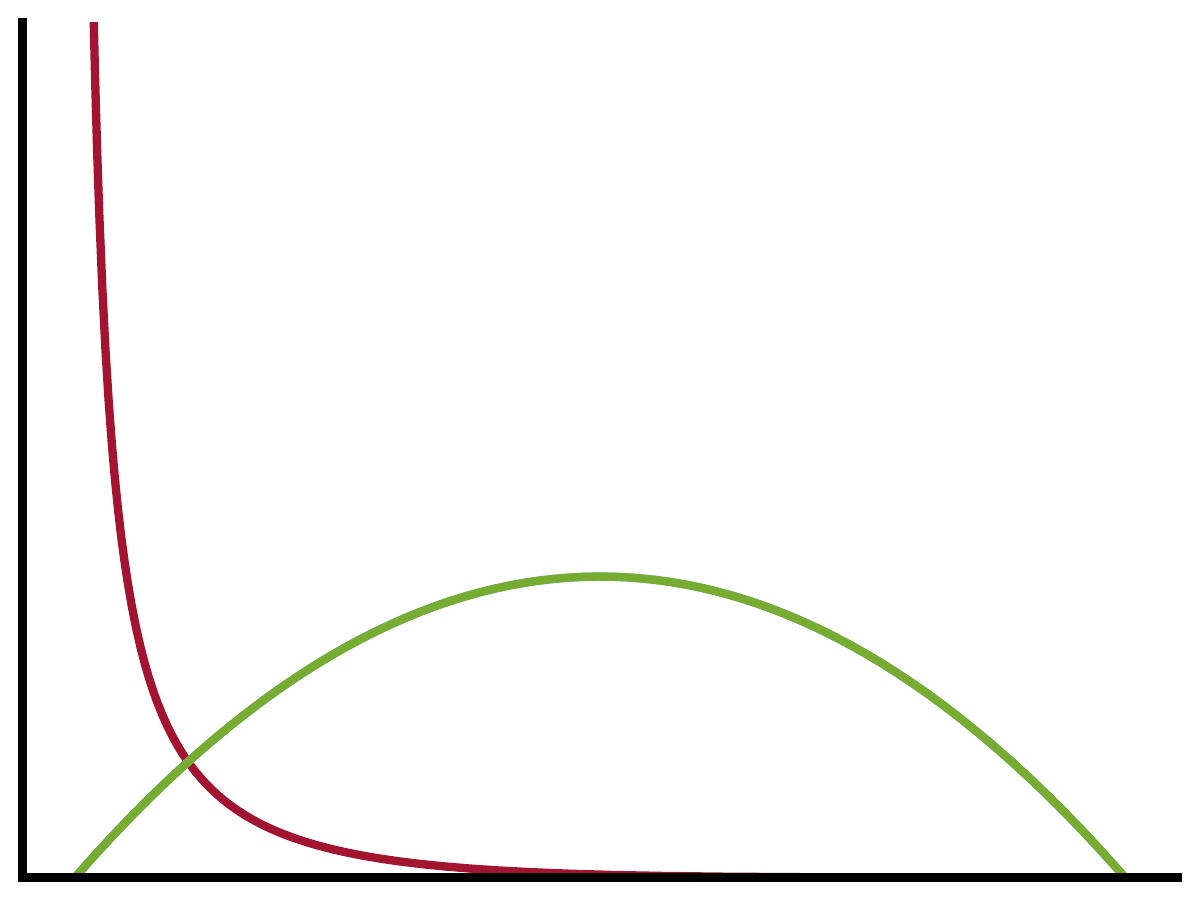}}}}&66.43/57.20&\textbf{41.45}/37.43&\textbf{15.78}/10.00&17.65/13.29&35.33/29.48\\

    \midrule
    \makecell[l]{
        $\textcolor{sparse}{\beta_s=\beta(4,3)}$\\
         $\textcolor{moderate}{\beta_m=\beta(2,2)}$ \\ 
         $\textcolor{dense}{\beta_d=\beta(2,3)}$ 
        } &\makecell{\raisebox{-0.1\height}{\resizebox{0.45\linewidth}{!}{\includegraphics{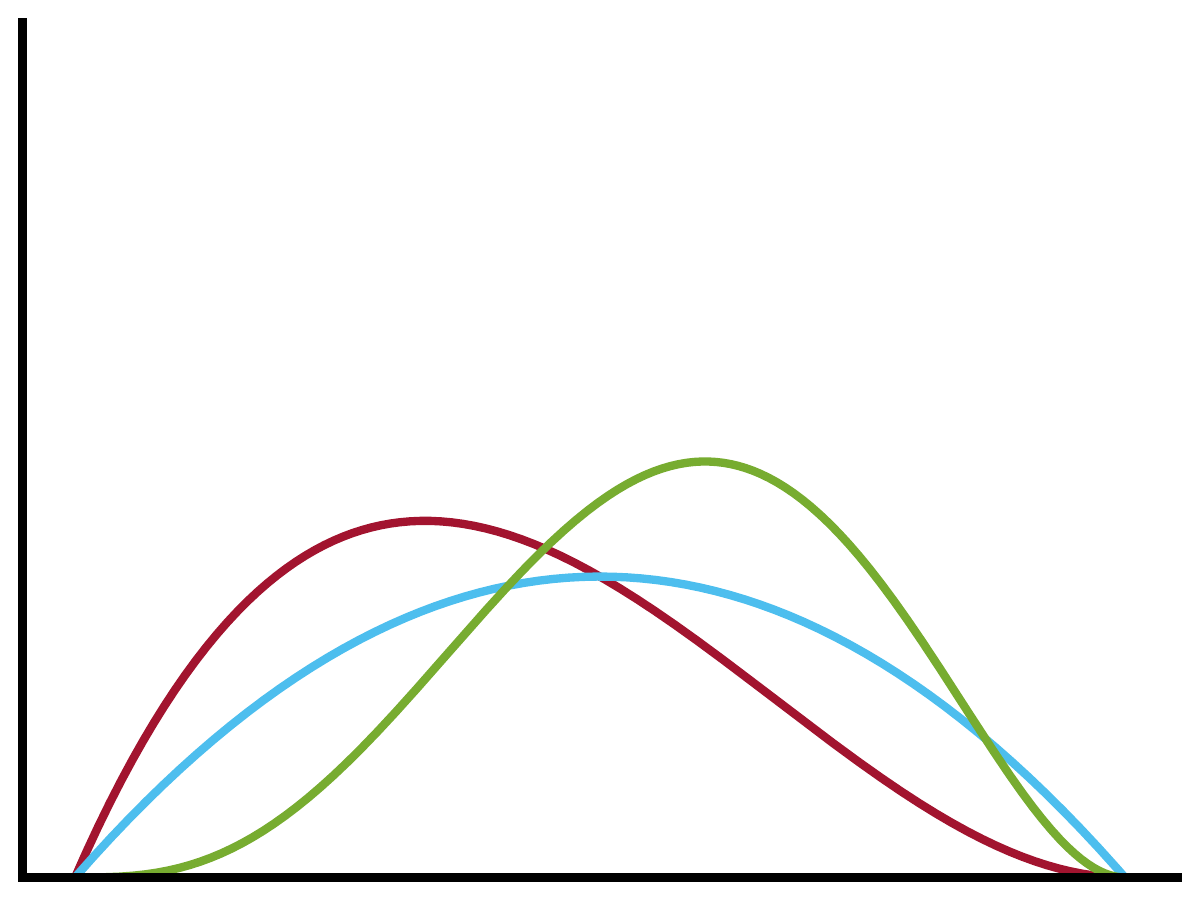}}}}&66.19/57.21&39.98/36.07&15.48/10.08&17.78/13.77&34.86/29.28\\
    \midrule
    \makecell[l]{
        $\textcolor{sparse}{\beta_s=\beta(2,2)}$\\
         $\textcolor{moderate}{\beta_m=\beta(2,3)}$ \\ 
         $\textcolor{dense}{\beta_d=\beta(0.1,5)}$ 
        } & \makecell{\raisebox{-0.1\height}{\resizebox{0.45\linewidth}{!}{\includegraphics{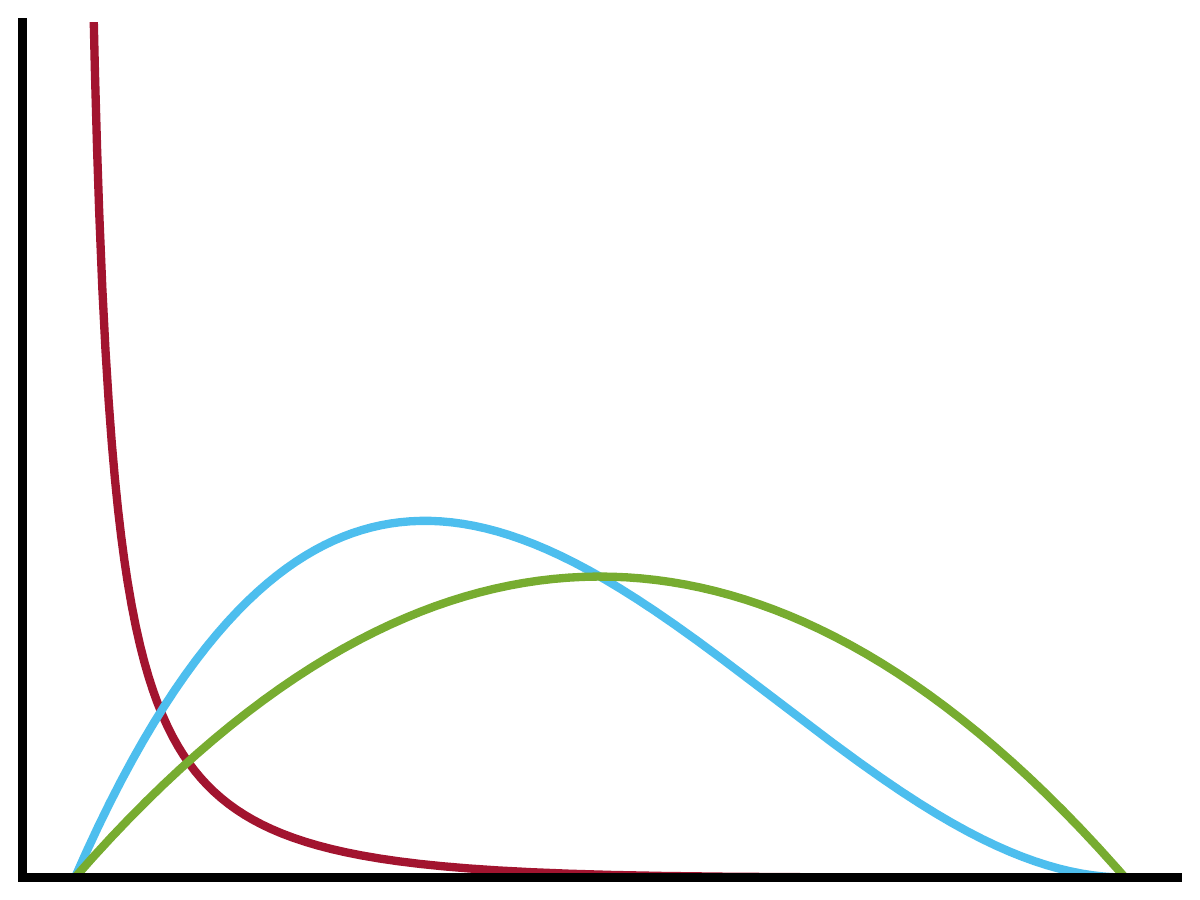}}}}&66.02/56.53&40.86/36.76&15.37/9.84&18.10/13.85&35.07/29.24\\
    \midrule
    \makecell[l]{
        $\textcolor{sparse}{\beta_s=\beta(10,1)}$\\
         $\textcolor{moderate}{\beta_m=\beta(2,2)}$ \\ 
         $\textcolor{dense}{\beta_d=\beta(0.1,5)}$ 
        }&\makecell{\raisebox{-0.1\height}{\resizebox{0.45\linewidth}{!}{\includegraphics{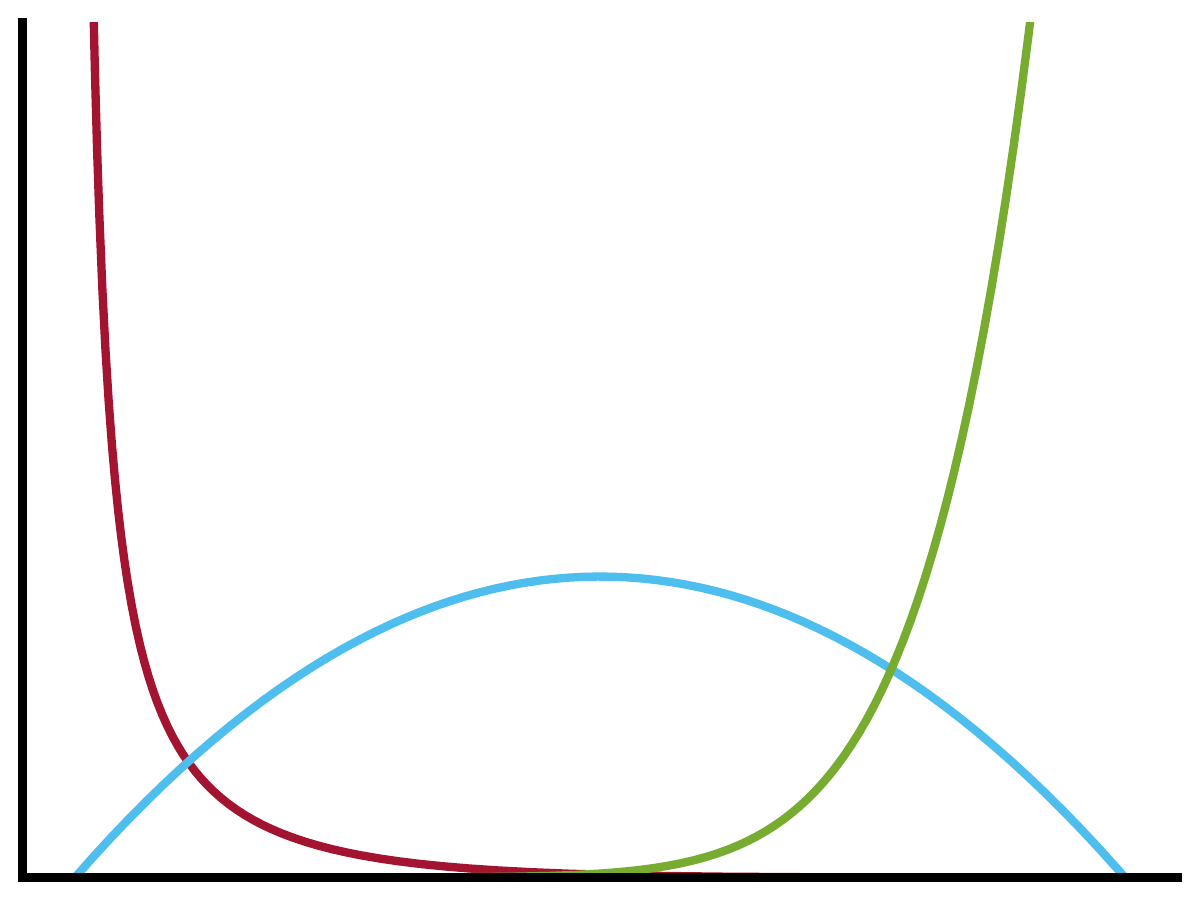}}}}&66.32/56.99&41.22/\textbf{37.88}&15.65/10.08&18.18/14.18&35.35/29.64\\
    \midrule
    \makecell[l]{
    $\textcolor{sparse}{\beta_s=\beta(0.1,5)}$\\
     $\textcolor{moderate}{\beta_m=\beta(2,2)}$ \\ 
     $\textcolor{dense}{\beta_d=\beta(10,1)}$ 
    }&\makecell{\raisebox{-0.1\height}{\resizebox{0.45\linewidth}{!}{\includegraphics{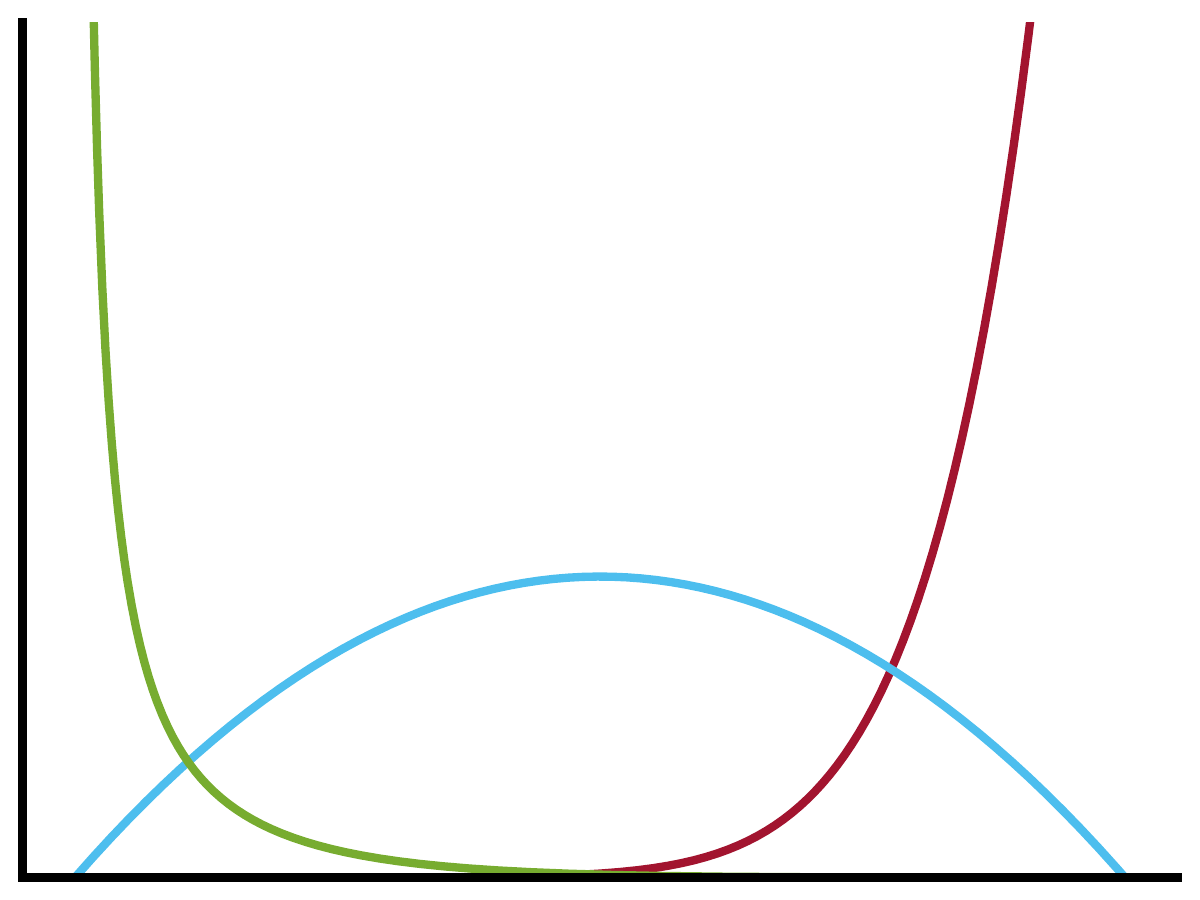}}}}&66.23/57.03&39.42/35.48&15.18/9.82&16.48/12.56&34.33/28.72\\
    \bottomrule
    \end{tabular}}
    \vspace{-1.5em}
\end{table}
\end{itemize}
\section{CONCLUSION}
In this paper, we provide a deep insight into the feasibility of applying MSDA techniques to radar point clouds. Additionally, we introduce a Class-Aware PillarMix method that merges pillar-based augmentation with MixUp strategies, specifically tailored to radar data. By assigning distinct $\beta$ distributions to each class, our approach increases the diversity of generated samples and significantly improves 3D object detection in low-data scenarios.

To the best of our knowledge, this work represents the first attempt to explore the application of MSDA to radar point clouds. We aim for this research to serve as a foundation and inspiration for future studies on MSDA in radar-based perception tasks. 
Moving forward, we endeavor to explore strategies for optimizing the $\beta$ parameters during training. We also plan to extend this approach to other contexts, including semi-supervised learning and domain adaptation.
\addtolength{\textheight}{-12cm}   




\bibliographystyle{IEEEtran}
\bibliography{IEEEabrv,ref}

\end{document}